\documentclass[preprint, 3p]{elsarticle}

\usepackage{multirow}
\usepackage{lineno,hyperref}
\modulolinenumbers[5]
\usepackage{mathtools}
\usepackage{amsmath}
\DeclareMathOperator{\sgn}{sgn}
\usepackage[mathscr]{euscript}
\usepackage{algorithm}
\usepackage{algpseudocode}
\usepackage{subfig}

\journal{Neurocomputing}
\begin{document}

\begin{frontmatter}

\title{Desire Backpropagation: A Lightweight Training Algorithm\\
for Multi-Layer Spiking Neural Networks\\
based on Spike-Timing-Dependent Plasticity}

\author[astar]{Daniel Gerlinghoff}
\author[astar]{Tao Luo\corref{mycorrespondingauthor}}
\cortext[mycorrespondingauthor]{Corresponding author}
\ead{leto.luo@gmail.com}
\author[astar]{Rick Siow Mong Goh}
\author[nus]{Weng-Fai Wong}

\address[astar]{Institute of High Performance Computing (IHPC), Agency for Science, Technology and Research (A*STAR), 1 Fusionopolis Way, \#16-16 Connexis, Singapore 138632, Republic of Singapore}
\address[nus]{Department of Computer Science, National University of Singapore, Computing 1, 13 Computing Drive, Singapore 117417, Republic of Singapore}

\begin{abstract}
Spiking neural networks (SNNs) are a viable alternative to conventional artificial neural networks when resource efficiency and computational complexity are of importance. A major advantage of SNNs is their binary information transfer through spike trains which eliminates multiplication operations. The training of SNNs has, however, been a challenge, since neuron models are non-differentiable and traditional gradient-based backpropagation algorithms cannot be applied directly. Furthermore, spike-timing-dependent plasticity (STDP), albeit being a spike-based learning rule, updates weights locally and does not optimize for the output error of the network. We present {\em desire backpropagation}, a method to derive the desired spike activity of all neurons, including the hidden ones, from the output error. By incorporating this desire value into the local STDP weight update, we can efficiently capture the neuron dynamics while minimizing the global error and attaining a high classification accuracy. That makes desire backpropagation a spike-based supervised learning rule. We trained three-layer networks to classify MNIST and Fashion-MNIST images and reached an accuracy of 98.41\% and 87.56\%, respectively. In addition, by eliminating a multiplication during the backward pass, we reduce computational complexity and balance arithmetic resources between forward and backward pass, making desire backpropagation a candidate for training on low-resource devices.
\end{abstract}


\end{frontmatter}

\section{Introduction}
Spiking Neural Networks (SNNs)~\cite{neil2016learning, farabet2012comparison, o2013real} have gained significant attention in recent years due to their potential for applications that require low-resource computing~\cite{lu2021autonomous}. One of the key advantages of SNNs is their highly efficient inference process, since it does not involve multiplication operations~\cite{lee2021accurate, blouw2019benchmarking, stromatias2015scalable}. This was further enabled by the development of custom hardware architectures specialized in the processing of spike information~\cite{yang2022coreset, gerlinghoff2022resource, luo2021nc, gerlinghoff2021e3ne, lin2020scalable, zhang2020low, aung2021deepfire, aung2023deepfire2}. However, supervised training of SNNs has remained a challenging problem. Due to the non-differentiable nature of neuron models, such as the leaky integrate-and-fire~\cite{koch1998methods} model, the standard backpropagation algorithm used for training traditional neural networks is not directly applicable.

For this reason, several supervised training algorithms for SNNs have been proposed in recent years. While each of them sets a specific focus, they can be broadly classified as follows:
\begin{enumerate}
    \item Convert the weights of a pre-trained traditional ANN to an SNN while considering the constraints and behaviour of the spiking neurons. Methods include fine-tuning of weights and normalization~\cite{rueckauer2017conversion, sengupta2019going}. While exhibiting great scalability, those methods do not factor in temporal spike information and are computationally expensive.
    \item Modify the standard backpropagation algorithm to enable gradient-based training of SNNs, for example by using approximations, surrogate gradients or stochastic models~\cite{thiele2019spikegrad, shrestha2018slayer, qiao2021direct, gardner2015learning}. Just like the first class of training methods, they do not accurately capture the neuron dynamics. The approximations might also lead to a degradation in accuracy.
    \item Directly adjust weights based on synaptic plasticity, considering pre- and post-synaptic spike events~\cite{feldman2012spike, ponulak2010supervised, mohemmed2012span}. Those methods are the most biologically plausible ones, but their locality restricts most of them to the training of only single-layer networks.
\end{enumerate}

There have been attempts of combining the second and third class of training algorithms to enable multi-layer networks to be trained using spike-timing-dependent plasticity (STDP)~\cite{tavanaei2019bp, shrestha2019approximating, liu2021sstdp}. This, however, leads to increased computational cost during the backward pass and weight update. In applications, where a resource-constrained device not only performs inference but also improves the SNN model, the number and complexity of operations matter.

To address the challenges in this scenario, we propose Desire Backpropagation, a lightweight training algorithm that involves backpropagating a ternary \textit{desire} value which serves as a spike target. Its ternary nature eliminates a multiplication operation in the backward pass, thereby achieving a balance in the number and type of operations between the forward and backward pass. Hence, an existing SNN inference device can be enabled to also carry out model training without additional hardware logic. The desire value is computed for neurons in both output and hidden layers, and is then used for weight updates following the STDP learning rule. STDP can be realized without a multiplication operation. Furthermore, it is local and takes into account the temporal spike information, which allows it to capture the underlying dynamics of the biological neuron model. Pairing STDP with desire backpropagation overcomes its restriction of learning only for a single layer.

We evaluate the performance of our proposed algorithm on the benchmark datasets MNIST \& Fashion-MNIST and compare it with state-of-the-art SNN training methods. Fashion-MNIST is a more challenging dataset, which demonstrates the the benefit of training multi-layer models and their ability to capture more complex patterns. Additionally, we show that Desire Backpropagation uses fewer and computationally less complex operations than conventional backpropagation.

The paper is organized as follows: Section~\ref{sec: related works} introduces existing spike-based learning algorithms in greater detail. We then explain relevant fundamental concepts on SNN architecture and training in Section~\ref{sec: background}. This is followed by a description of our proposed learning rule in Section~\ref{sec: proposed learning procedure}, distinguishing the computations applied to output and hidden layers. The results of our experiments and the insights we gained during the training process are presented in Section~\ref{sec: experiment}. Lastly, we draw a conclusion. The source code for this paper is available on GitHub\footnote{\url{https://github.com/DanielGerlinghoff/desire-backpropagation}}.

\section{Related research} \label{sec: related works}
The learning rules of spiking neural networks are inspired by the inner workings of biological brains. In order for a neuron model and learning rule to be deemed biologically plausible, training and inference should be spike-based, i.e., operate on the temporal information of the spike train and interface using spike information~\cite{hao2020biologically}. Spike-timing-dependent plasticity is such a rule, that only relies on locally available spike information, and was used for unsupervised learning~\cite{feldman2012spike}. In addition to adjusting weights purely based on spike timings, regularization was applied to spikes and/or weights to improve the performance and robustness. Learning neurons in competition through a winner-takes-all mechanism that inhibits the spiking of laterally adjacent neurons to prevent them from learning the same patterns~\cite{masquelier2010learning}. Diehl~et~al.~\cite{diehl2015unsupervised} furthermore offset the spike trace by a negative constant and used exponential weight decrease to increasingly "disconnect" rarely spiking neurons. An adaptive membrane threshold balances the firing activity across neurons and prevents single neurons from dominating the output. Kheradpisheh~et~al.~\cite{kheradpisheh2018stdp} coupled an unsupervised STDP-based feature extraction network with a linear SVM classifier to enable supervised learning. Similarly, Thiele~et~al.~\cite{thiele2018timescale} uses a custom event-based classifier to allow the learning to depend on the output error.

To use the STDP rule directly for supervised learning, it was separated into the STDP and anti-STDP processes, which refer to weight strengthening and weakening, respectively~\cite{ponulak2010supervised, sporea2013supervised, wang2014online}. Ponulak~et~al.~\cite{ponulak2010supervised} presented ReSuMe, which trains a network for a timing-accurate spike sequence. The STDP process is applied whenever a spike in the desired sequence occurs, while anti-STDP is applied for every actual output spike. Once the actual spike train matches the desired one, STDP and anti-STDP process counteract each other and weights are not modified. A \textit{learning window} is used for the spike timing to scale the magnitude of the weight update. Multiple variants of ReSuMe have been developed, which are either incorporating an additional synaptic delay (DL-ReSuMe~\cite{taherkhani2015dl}) or using triplet-STDP (T-ReSuMe~\cite{lin2016improved}).
ReSuMe was also adopted by Wang~et~al.~\cite{wang2014online}. Weights of their output layer are updated at the time of actual output spikes only. The sign of the weight update is determined by whether or not the output neuron is desired to spike. The idea of spiking desire was used earlier by the Tempotron~\cite{gutig2006tempotron} to determine the sign of the weight update, while the magnitude is proportional to the output error. In a similar manner, PBSNLR~\cite{xu2013new} tries to match the output spike timings with a desired spike train by adjusting weights, whenever there is a spike at an undesired time or a failure to spike at a desired time step. Those rules are only applied to a single-layer MLP or the output layer of the SNN as hidden neurons lack the information about spiking desire.

Other spike-based supervised learning algorithms are based on the Widrow-Hoff rule, e.g. SPAN~\cite{mohemmed2012span, mohemmed2013training}. It adjusts the neurons of the output layer according to the contribution to their respective output errors. Transforming the spike train into a continuous function by convolving it with a kernel renders it more amenable to mathematical operations. This rule is, again, applied to the last layer only due to the ability to directly compute the output error. Tavanaei \& Maida~\cite{tavanaei2019bp} improved the Widrow-Hoff rule to train the last two layers of a network.

Seeing the benefits of spike-based learning, SpikeProp~\cite{bohte2000spikeprop} adopted traditional backpropagation to SNN neuron models through approximation of the threshold function of the neuron model. Several derivations of this work use momentum or add trainable parameters such as synaptic delay or firing threshold~\cite{luo2017extended, matsuda2016bpspike}. But encoding information using only a single spike limits their effectiveness in solving complex spatiotemporal problems. Multi-ReSuMe~\cite{sporea2013supervised} adopts the original training method to support multiple spikes and multiple layers by backpropagating the output error. Shrestha~et~al.~\cite{shrestha2019approximating} uses error neurons to record local errors during the forward pass. They exhibit the same behaviour as forward neurons, which effectively doubles the neuron count and requires additional resources. Mirsadeghi~et~al.~\cite{mirsadeghi2021stidi} approximate error backpropagation in SNN and execute the weight adjustment based on STDP principles. Liu~et~al.~\cite{liu2021sstdp} approximate partial derivatives of the backpropagation with an STDP-like expression, enabling the error to be propagated back through the network. Luo~et~al.~\cite{luo2022supervised} combined temporal spike information with backpropagation by looking at the first occurrence of an erroneous spike in the spike train. The direction of the weight update is determined by the type of error.

Avoiding backpropagation through the SNN altogether, recent works have experimented with global error signals. GLSNN~\cite{zhao2020glsnn} computes the output errors, which serve as input to a separate set of feedback layers, yielding the local spike targets. This form of feedback alignment does not, however, consider the temporal relation between spikes when updating the weights. DECOLLE~\cite{kaiser2020synaptic} uses readouts at every layer to supply them with auxiliary targets. Those targets either need to be handcrafted or, in practise, are the same as the final output target. The latter approach has been uses by e-prop~\cite{bellec2020solution} to eliminate the temporal assignment problem in recurrent SNN.

Learning of spiking neural networks has also been adopted from conventional artificial neural networks. However, learning rules do often not adhere to the characteristic of biological plausibility mentioned earlier. Wu~et~al.~\cite{wu2019deep} apply error backpropagation to SNNs by updating the synaptic weights based on the total number of output spikes rather than spike timing.

\section{Background} \label{sec: background}
Like traditional artificial neural networks, SNNs have a layered structure with each layer consisting of an array of neurons. Adjacent neurons are connected via synapses, which can be understood as neuronal junctions to transfer spikes. The two neurons connected by a synapse are referred to as pre-synaptic and post-synaptic neuron, respectively. In this section, we consider a generic fully-connected layer. Post-synaptic neurons of this layer are indexed with $o$. They receive spikes from pre-synaptic neurons $i$.

\subsection{Neuron model} \label{sec: neuron model}
Neuron models are a mathematical description of the neuron behavior observed in biological experiments. Both the Hodgkin-Huxley model~\cite{hodgkin1952quantitative} and the Izhikevich model~\cite{izhikevich2003simple} are accurate representations, but are computationally expensive when used for large networks consisting of thousands of neurons. The Spike Response Model (SRM)~\cite{gerstner1993spikes} simulates the neuron's response to a spike using filter kernels. The SRM is a generalization of the leaky integrate-and-fire (LIF) model~\cite{koch1998methods}. Both are biologically plausible, albeit simpler than the first two models. For that reason, they are commonly used for efficient SNN implementations~\cite{fang2020encoding, cassidy2013cognitive}. Our experiments are also based on LIF neurons.

The information transfer between neurons comprises binary spike events, which can occur at any time step $t \in [0, T-1]$, with $T$ denoting the length of the spike trains. A LIF neuron $o$ retains its internal state, i.e. its membrane potential $p_o$, between the time steps. Additionally, a leak term $\beta_p \in [0, 1]$ causes a decay of the membrane potential over time. An input spike $s_i$ causes $p_o$ to be increased by weight $w_{oi}$, where $w_{oi}$ can be a signed value. If the membrane potential surpasses a threshold $\theta_p$, an output spike $s_o$ is fired and the membrane potential is reset. Due to the non-linearity inherent in the neuron model, no additional activation function is needed at the neuron output. The behavior of the LIF neuron can be illustrated as in Figure~\ref{fig: lif neuron}, and expressed by Equation~\ref{eqn: lif neuron}.

\begin{equation} \label{eqn: lif neuron}
    p_o[t+1] = \beta_p p_o[t] + \sum_i w_{oi} s_i[t] - \theta_p s_o[t]
\end{equation}

\subsection{Spike encoding}
Various methods exist to encode integer or floating-point values into spike trains~\cite{kempter1999spike, borst1999information, gutig2014spike, tang2020rank, sboev2020solving, wang2022efficient}. In this work, we use the firing-rate encoding as detected in the brain by Zhang~\& Linden~\cite{zhang2003other}, where the number of spikes within the fixed-length spike train is proportional to the real value they represent. That is opposed to temporal encoding, where the precise timing of the spike carries information.

Our rate-encoded spike trains are generated at the input layer by exploiting the dynamics of an integrate-and-fire neuron (see Section~\ref{sec: spike generation} for more details). The encoded information propagate through the network via spikes, with decoding to real values only occurring when evaluating the loss function after the output layer has completed or when computing local errors during backpropagation. Our weight update mechanism itself does not require decoding and acts purely upon spike occurrences.

\subsection{Spike-timing-dependent plasticity} \label{sec: spike-timing-dependent plasticity}
\begin{figure}[t]
    \centering
    \begin{minipage}[t]{0.48\textwidth}
        \centering
        \includegraphics[width=0.84\textwidth]{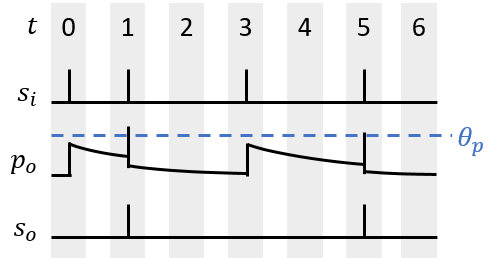}
        \caption{Input spikes $s_i$ and membrane potential $p_o$ of LIF neuron $o$ for every time step $t \in [0, 6]$. If threshold $\theta_p$ is exceeded, output spikes $s_o$ are generated.}
        \label{fig: lif neuron}
    \end{minipage}
    \hfill
    \begin{minipage}[t]{0.48\textwidth}
        \centering
        \includegraphics[width=0.8\textwidth]{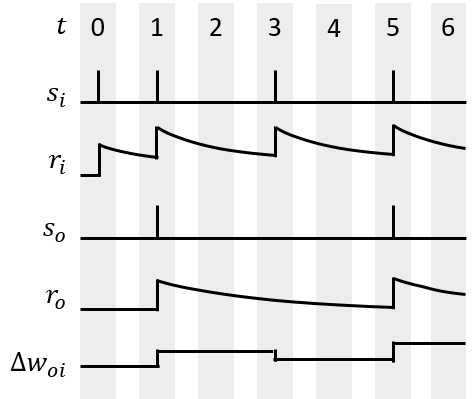}
        \caption{Accumulated weight adjustment $\Delta w_{oi}$ using classical STDP, which operates on input spikes $s_i$ and output spikes $s_o$ with respective traces $r$.}
        \label{fig: stdp}
    \end{minipage}
\end{figure}

According to Hebb~\cite{hebb1949organization}, a weight between two neurons is influenced by the spiking activity of both these neurons. This led to the development of the spike-timing-dependent plasticity (STDP) rule~\cite{gerstner1996neuronal, caporale2008spike}, which is a purely spike-based method of training spiking neural networks. It follows the idea of \textit{neurons that fire together wire together}~\cite[p.~64]{shatz1992developing}. Synaptic weight updates are dependent on relative differences between the spike timing of pre- and post-synaptic neurons. A pre-synaptic spike occurring before a post-synaptic spike is an indication that the pre-synaptic neuron $i$ contributes to the firing of the post-synaptic neuron $o$. In this case, the connection between $i$ and $o$ is strengthened by increasing weight $w_{oi}$. And vice versa, if pre-synaptic neuron $i$ fires before post-synaptic neuron $o$, the weight connecting them is decreased. The update of weight $w_{oi}$ at a time step $t$ can be expressed as:

\begin{equation} \label{eqn: stdp}
    \Delta w_{oi}[t] = \eta^+ s_o[t] r_i[t] - \eta^- s_i[t] r_o[t].
\end{equation}

\noindent
where $s_i$ and $s_o$ are binary spike trains of input and output neuron, respectively. $\eta^+$ and $\eta^-$ are learning rates. Decaying spike traces $r_i$ and $r_o$ account for the time delay between pre- and post-synaptic spikes. The magnitude of the weight update increases as the time period between pre- and post-synaptic spikes becomes shorter. Spike traces are generated by first decaying any existing trace value with an exponential kernel. After that, the trace is incremented in case the neuron fires a spike (see Equation~\ref{eqn: traces}). The decay rate $\beta_r < 1$ is a hyper-parameter, which can be trimmed to adjust how the delay between spikes influences the weight update. Figure~\ref{fig: stdp} shows the dynamics of spike traces and weight update expressed by Equation~\ref{eqn: stdp}.

\begin{equation} \label{eqn: traces}
    r_{\{o|i\}}[t] = \beta_r * r_{\{o|i\}}[t-1]+ s_{\{o|i\}}[t]
\end{equation}

\section{Proposed learning rule} \label{sec: proposed learning procedure}
Spike-timing-dependent plasticity is an effective way of adjusting the relationship between neurons in a way that increases their sensitivity to important features present in the dataset~\cite{fu2021ensemble}. The weight updates are, however, required to be guided, if one wishes to perform supervised learning. In that case, besides extracting features, neuron weights are tuned for reduction of the network's output error. Each neuron in the network is assigned a {\em desire} value, which is derived from the training labels and determines the direction of the weight update. Through local loss computation, desire values can be assigned to the neurons in the hidden layers.

Since we adopt the backpropagation algorithm, the training routine follows that of traditional deep neural networks: During the forward pass, a sample of the training dataset, which is encoded as spike trains, generate spikes and spike traces in the hidden and output neurons. The backward pass employs the desire backpropagation algorithm to derive desire values from training labels and output error. Lastly, the weights are updated following the STDP rule, utilizing spike trains/traces and being guided by desire values.

\subsection{Desire-based weight update} \label{sec: weight update}
The original STDP rule updates the weight $w_{oi}$ whenever there is a spike in either neuron $i$ or neuron $o$. For supervised learning, however, weight updates should be tied to the activity of the post-synaptic neuron~$o$ only, as those are used to compute the subsequent layer and ultimately lead to the output result. Our weight update therefore only takes the post-synaptic spike train $s_o$ and pre-synaptic trace $r_i$ into account. Weights between neurons with higher activity experience larger adjustments, following the idea of Hebbian learning. This is expressed in Equation~\ref{eqn: desire weight update}, which applies equally to all weights in the network.

\begin{equation} \label{eqn: desire weight update}
    \Delta w_{oi}[t] = d_o * \eta s_o[t] r_i[t].
\end{equation}

The STDP term of the update equation is not sufficient for supervised learning. That is, instead, accomplished by the desire term $d_o$, which serves two purposes. Firstly, it decides whether the weight is increased or decreased. Secondly, the weight update is masked if the neuron's spikes have an insignificant influence on the successive layer. Therefore, we define a ternary value $d_o \in \{-1, 0, +1\}$ with the three states describing that
\begin{itemize}
    \item $+1$: the neuron $o$ ought to spike and its weights $w_{oi}$ should increase for all $i$,
    \item $-1$: the neuron $o$ should not spike and its weights should decrease, or
    \item $0$: the activity of neuron $o$ is indifferent and the weights should not change
\end{itemize}

Besides disabling the weight update, a desire of $0$ will cause the neuron to not be considered during backpropagation, as will be seen in the next section. The proposed learning rule incorporates STDP, while the desire value depends on the output error. Therefore, it is considered a spike-based supervised learning rule.

\subsection{Desire backpropagation} \label{sec: desire backpropagation}
\begin{figure}[t]
    \centerline{\includegraphics[width=0.75\textwidth]{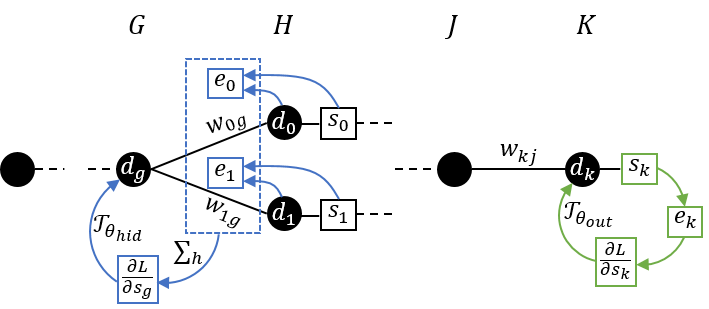}}
    \caption{Minimal neural network, which only shows two arbitrary hidden layers $G$ \& $H$ and the last two layers $J$ \& $K$. The steps of computing output desire $d_k$ and hidden desire $d_g$ are visualized by the green and blue arrows, respectively.}
    \label{fig: network architecture}
\end{figure}

So far, we have used indices $i$ and $o$ to index pre- and post-synaptic neurons of a generic linear layer. In the following, however, there is a differentiation between the output layer and the hidden layers of the network. Figure~\ref{fig: network architecture} sketches a network with two subsequent hidden layers $G$ and $H$, and the last two layers $J$ and $K$, each with a minimal number of neurons for the sake of readability. We replace indices $o$ and $i$ with the respective layer descriptors. The figure gives an overview of the steps involved in the computation of the desire values $d_k$ and $d_g$, based on the output error and local errors, respectively.

\subsubsection{Output layer}
Supervised learning algorithms aim to reduce the output error of the last network layer. In rate-encoded SNNs, each of the network's output neurons $k$ generates a spike sequence $s_k$ with $T$ binary elements. Accumulation over that sequence yields the spike count. The output error $e_k$ of a neuron is the difference between the normalized spike count and a target $\hat{s_k}$, given in Equation~\ref{eqn: desire error output}. The target is derived from the training labels. If neuron $k$ represents the target class, it should fire $T$ spikes, and remain silent otherwise. That is formulated in Equation~\ref{eqn: desire error output}.

\begin{equation} \label{eqn: desire error output}
    e_k = \frac{1}{T} \sum_{t=0}^{T-1} s_k[t] - \hat{s_k}
    \quad \text{with} \quad
    \hat{s_k} =
    \begin{cases}
        1, & \text{if } k \text{ is target class} \\
        0, & \text{otherwise}
    \end{cases}
\end{equation}

To quantify the performance of the network we apply the squared error loss function in Equation~\ref{eqn: desire loss} to the output error $e_k$. The backpropagation algorithm requires the derivative of the loss function with respect to a single output spike train $s_k$, which resolves to Equation~\ref{eqn: desire loss derivative} after applying the chain rule.

\begin{equation} \label{eqn: desire loss}
    L = \frac{1}{2} \sum_k e_k^2
\end{equation}

\begin{equation} \label{eqn: desire loss derivative}
    \frac{\partial L}{\partial s_k} = e_k
\end{equation}

The update of weight $w_{kj}$ is guided by a desire value $d_k$ according to Equation~\ref{eqn: desire weight update}. Due to the structural similarity between feed-forward ANN and SNN, we derive the desire by applying gradient decent. The gradient of the loss function with respect to the weight is given in Equation~\ref{eqn: desire backpropagation 1}, where variables are denoted using the SNN symbols from the previous sections. Due to the non-differential LIF neuron model, we follow~\cite{shrestha2019approximating} and approximate the second term with a straight-through estimator, which has a gradient of one. The first term describes the impact of the weight on the membrane potential $p_k$. The binary nature of the spikes causes the gradient to be 1, resulting in Equation~\ref{eqn: desire backpropagation 2}.

\begin{equation} \label{eqn: desire backpropagation 1}
    \frac{\partial L}{\partial w_{kj}} =
    \frac{\partial p_k}{\partial w_{kj}}
    \frac{\partial s_k}{\partial p_k}
    \frac{\partial L}{\partial s_k} 
\end{equation}

\begin{equation} \label{eqn: desire backpropagation 2}
    \frac{\partial L}{\partial w_{kj}} =
    1 * 1 * \frac{\partial L}{\partial s_k} =
    e_k
\end{equation}

By following the descending gradient of the loss function, we obtain an expression for the desired change in spiking activity, which leads us to our desire value $d_k$. Since the magnitude of the weight update is determined by the spike timings using STDP, we apply the ternarization function $\mathscr{T}_\theta$ (see Equation~\ref{eqn: desire ternarization}). $\sgn(\cdot)$ is the sign function, which returns $-1$ for negative and $+1$ for positive values. $\theta$ is a threshold value to stabilize the training for small errors, which we call $\theta_{\text{out}}$ for the output layer in particular. That yields the final expression in Equation~\ref{eqn: desire value output}. In case of the output layer, only weights connecting to the target neuron can get increased, while all others might get decreased.

\begin{equation} \label{eqn: desire ternarization}
    \mathscr{T}_\theta(x) =
    \begin{cases}
        \sgn x, & \text{if } |x| > \theta \\
        0,      & \text{otherwise}
    \end{cases}
\end{equation}

\begin{equation} \label{eqn: desire value output}
    d_k =
    \mathscr{T}_{\theta_{\text{out}}} \left(-\frac{\partial L}{\partial w_{kj}} \right) =
    \mathscr{T}_{\theta_{\text{out}}} \left(-e_k \right) =
    \begin{cases}
        -\sgn e_k, & \text{if } |e_k| > \theta_{\text{out}} \\
        0, & \text{otherwise}
    \end{cases}
\end{equation}

\subsubsection{Hidden layers}
Desire backpropagation enables the training of multi-layer neural networks. In contrast to previous methods, such as~\cite{tavanaei2019bp}, where the output error of the last layer is backpropagated, we compute the error locally. That allows us to train networks with a higher number of layers. As shown in Figure~\ref{fig: network architecture}, desire values $d_g$ of hidden layer $G$ are determined based on the desire values and spike trains of layer $H$.

In the previous section, we derived the dependency of a desire value on the loss gradient with regards to the spike train. To determine this gradient for a hidden neuron, we use the classical backpropagation as a starting point (see Equation~\ref{eqn: desire backpropagation 3}). Just as before, the straight through estimator and the squared error loss is applied to obtain Equation~\ref{eqn: desire backpropagation 4}.

\begin{equation} \label{eqn: desire backpropagation 3}
    \frac{\partial L}{\partial s_{g}} = \sum_h
    \frac{\partial p_h}{\partial s_g}
    \frac{\partial s_h}{\partial p_h}
    \frac{\partial L}{\partial s_h} 
\end{equation}

\begin{equation} \label{eqn: desire backpropagation 4}
    \frac{\partial L}{\partial s_{g}} = \sum_h
    w_{hg} * e_h
\end{equation}

To calculate the loss locally for layer $H$, the error is only dependent on the spike output and the desire values of this layer. Equation~\ref{eqn: desire error hidden} transforms the ternary desire values $d_h$ into targets of $\{0, 1\}$ for the normalized spike count. If the desire value is zero, however, an error cannot be calculated due to the indifference in spiking desire of the neuron. The loss is obtained by squaring the error as done before in Equation~\ref{eqn: desire loss}.

\begin{equation} \label{eqn: desire error hidden}
    e_h =
    \begin{cases}
        \frac{1}{T} \sum_{t=0}^{T-1} s_h[t] - (d_h + 1) / 2, & \text{if } d_h \neq 0 \\
        0, & \text{otherwise}
    \end{cases}
\end{equation}

The desire values of the preceding hidden layer $G$ are obtained by supplying the negative gradient to the ternarization function $\mathscr{T}_\theta$ with a threshold $\theta_{\text{hid}}$ specific for hidden layers. The final expression of a desire value $d_g$ is given in Equation~\ref{eqn: desire value hidden}. In summary, the computation only relies on desire values, spike trains and weights of layer $H$.

\begin{equation} \label{eqn: desire value hidden}
    d_g =
    \mathscr{T}_{\theta_{\text{hid}}} \left(-\frac{\partial L}{\partial s_g} \right) =
    \begin{cases}
        -\sgn \left( \sum_h w_{hg} e_h \right), & \text{if } \left| \sum_h w_{hg} e_h \right| > \theta_{\text{hid}} \\
        0, & \text{otherwise}
    \end{cases}
\end{equation}

Using Equation~\ref{eqn: desire value hidden}, the desire values of layer $G$ are set such that the desire in layer $H$ is satisfied. Although a pre-synaptic neuron $g$ cannot fulfill the desire of all post-synaptic neurons $d_h$, the sum over all $w_{hg} e_h$ gives an indication about the preference in the spiking activity of $g$. If neuron $g$ does not significantly contribute to the average fulfillment of the desire in layer $H$, no adjustment is made to its weights. This is reflected by a desire $d_g$ of zero.

If the network has been trained and has reached an equilibrium, the local errors are small enough to draw desire values to zero. In that case, weights theoretically stabilize and are not adjusted. The network's ability to generalize depends, however, on the number of neurons and layers. In practise, the output error will not be zero for all samples, which leads to continuing weight adjustments even towards the end of the training.

\subsection{Dropout} \label{sec: dropout}
After many training epochs, the accuracy of classifying the training dataset usually increases, while the accuracy on the validation dataset might remain unchanged or even decrease. This is the phenomenon of overfitting.  Dropout~\cite{srivastava2014dropout} is a technique to increase the generality of a network model, and has been shown to be effective for SNNs~\cite{lee2020enabling}.

Dropout simulates the training of multiple neural networks by randomly excluding neurons of each layer during the training phase. During the evaluation phase, dropout is not applied, which is equivalent to combining the results of all those neural networks. The parameter $P_{\text{drop}}$ gives the probability of a neuron to be dropped out. Since the neuron configuration needs to remain constant for all time steps, a binary dropout mask $m$ is generated for each layer at the beginning of each sample. An input neuron $i$, which is silenced by $m_i = 0$, does not contribute to the desire backpropagation and does not get its weights updated.

Dropout leads to a reduction of the number of spikes arriving at the next layer. To compensate for that, the influence of the active neurons on the membrane potential is increased by a factor $1 / (1 - P_{\text{drop}})$. Equation \ref{eqn: lif neuron} for the training phase is modified to include dropout as follows:

\begin{equation} \label{eqn: dropout}
    p_o[t+1] = \beta_p p_o[t] + w_{oi} s_i[t] * \frac{m_i} {1 - P_{\text{drop}}} - s_o[t].
\end{equation}

\section{Experiments and results} \label{sec: experiment}
\subsection{Spike generation} \label{sec: spike generation}
For the sake of comparability, we used popular image datasets for classification. Since those datasets are only available as two-dimensional pixel data, spike trains with additional timing information need to be generated. We use rate encoding, which conveys information through the quantity of spikes present in the spike train. We do not carry out any data preprocessing or augmentation.

Our input layer contains integrate-and-fire neurons (see Section~\ref{sec: neuron model}), whose synaptic current is equal to the the brightness value of the input $\in [0, 1]$. The membrane potential $p_o$ is not decayed, i.e. $\beta_p = 1$, and incremented by the constant input value during every time step. A brighter pixel leads to larger increments of $p_o$. Hence, the threshold $\theta_p$ is reached faster and more often and more spikes are generated. Therefore, the number of spikes in the spike train is proportional to the brightness value of the input pixel. The dropout rate $P_{drop}$ can be adjusted separately for the input layer.

\subsection{MNIST classification} \label{sec: MNIST classification}
\begin{figure}[t]
    \centering
    \begin{minipage}[t]{0.49\textwidth}
        \centering
        \includegraphics[width=\textwidth]{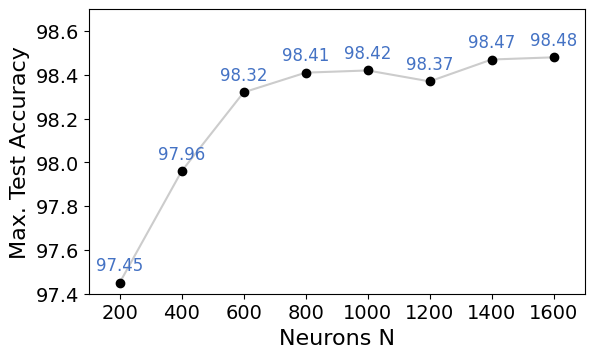}
        \caption{Classification accuracy (in \%) plotted against the number of hidden neurons $N$ in a 3-layer MLP architected as 784 -- $2N$ -- $N$ -- 10 and $L=20$.}
        \label{fig: plot architecture}
    \end{minipage}
    \hfill
    \begin{minipage}[t]{0.49\textwidth}
        \centering
        \includegraphics[width=\textwidth]{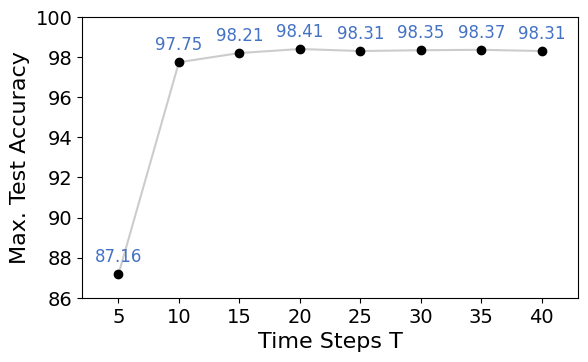}
        \caption{Accuracy (in \%) versus spike train length / number of time steps $T$ with $N$ set to 800.}
        \label{fig: plot timesteps}
    \end{minipage}
\end{figure}

\begin{figure}[t]
    \centerline{\includegraphics[width=0.50\textwidth]{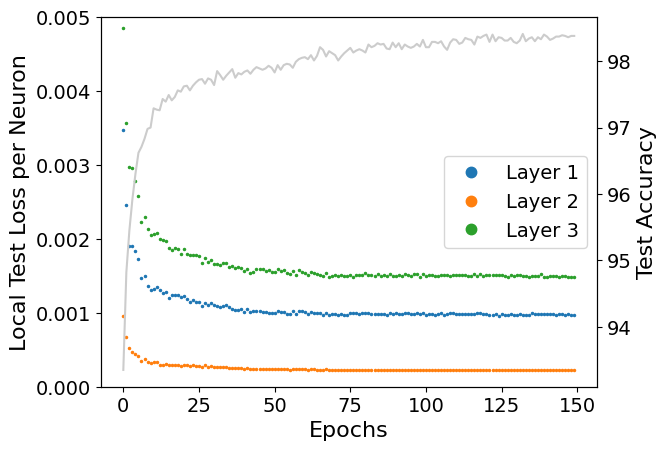}}
    \caption{Average local test losses for each neuron in layer 1, 2 and 3 throughout the training with 150 epochs. Accuracy on test dataset for each epoch (gray line).}
    \label{fig: local errors}
\end{figure}

\begin{figure}[t]
    \centerline{\includegraphics[width=1.0\textwidth]{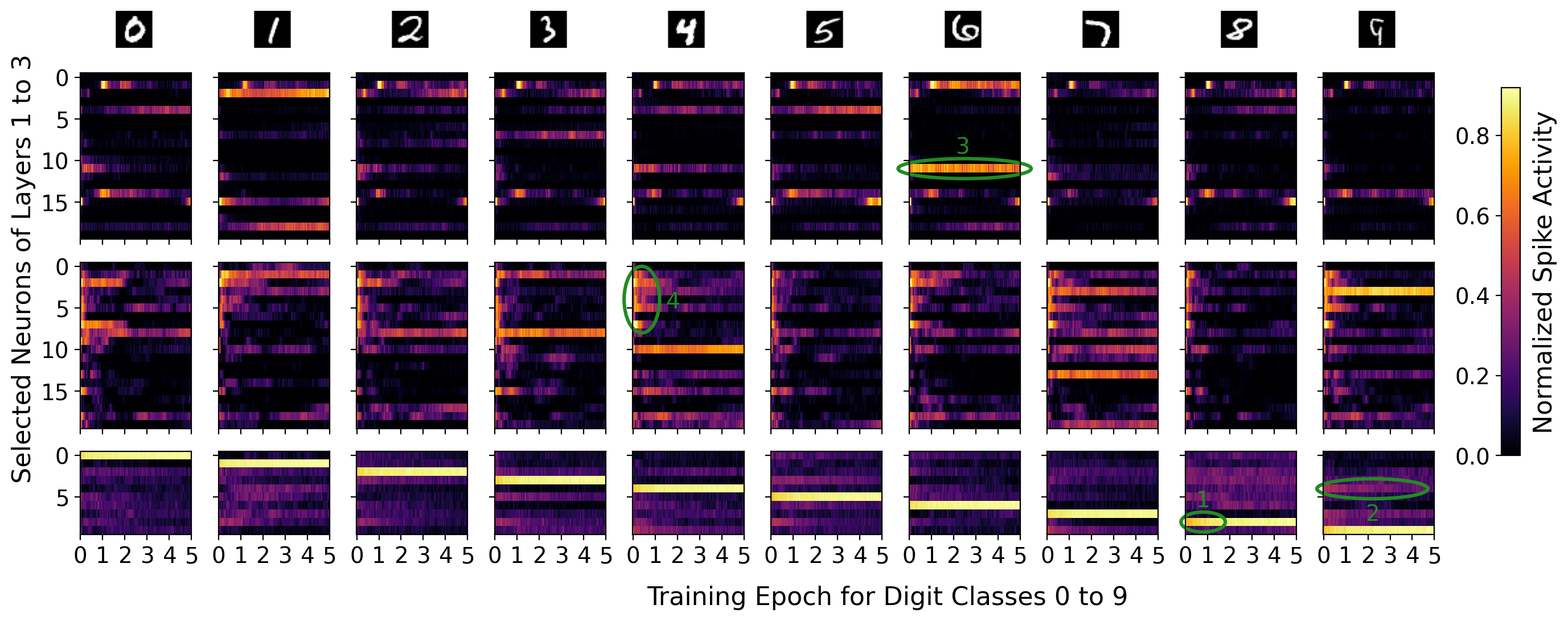}}
    \caption{Color-coded spike activity of neurons in the three layers over five training epochs. Plots are separated into columns for each of the ten handwritten digits 0 to 9. Four regions of interest are marked by green circles.}
    \label{fig: spike activity}
\end{figure}

\begin{figure}[t]
    \centerline{\includegraphics[width=1.0\textwidth]{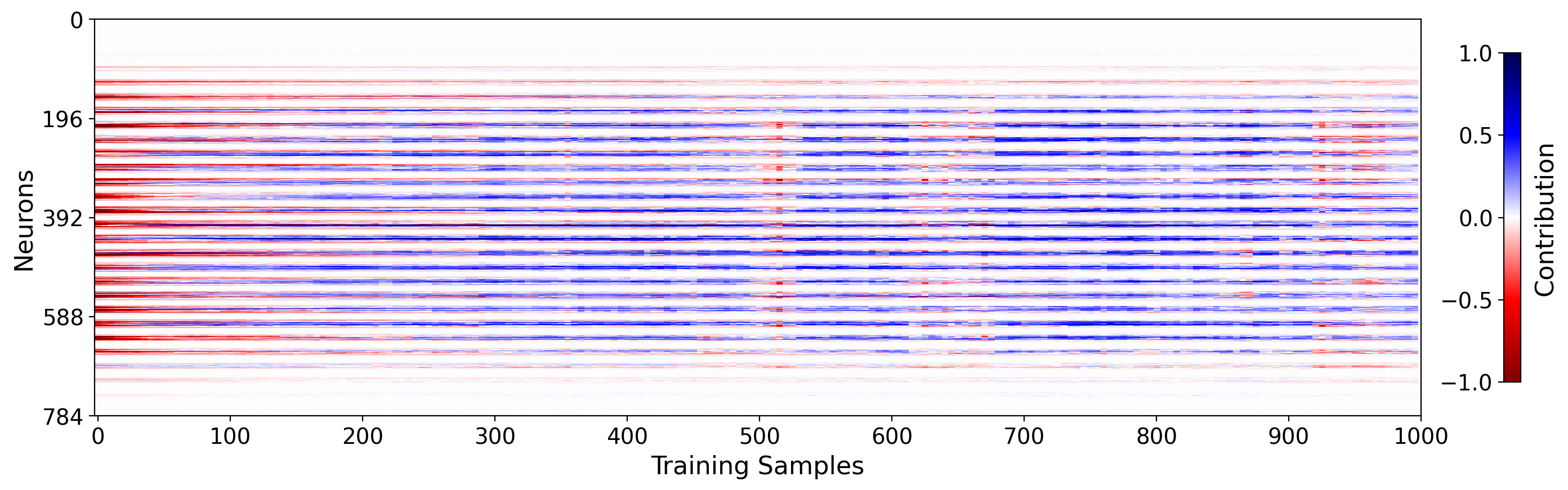}}
    \caption{Contribution of each of the input layer neurons to satisfy on average the desire of the neurons in the next layer, measured for the first 1000 samples in the first training epoch. White color represents no contribution, while the shades of red and blue stand for negative and positive contribution, respectively.}
    \label{fig: contribution}
\end{figure}

We trained a multilayer perceptron (MLP) on 60,000 training images of the handwritten digits dataset MNIST~\cite{lecun1998mnist}. The accuracy values in this section were obtained from the 10,000 test images. The network consists of three fully-connected layers with a variable neuron count of 784 -- $2 N$ -- $N$ -- 10. We use a variable number of time steps $T$ for the spike trains. Our PyTorch implementation of the spiking neural network allows an exponential decay of the learning rate. We start off with a learning rate $\eta = 1*10^{-5}$, which is reduced by 4\% at the end of every training epoch. Training is carried out for a total of 150 epochs. Other hyper-parameters include the firing threshold $\theta_p = 1.0$, the desire thresholds $\theta_{\text{hid}} = 0.05$ and $\theta_{\text{out}} = 0.30$, and the dropout rate $P_{\text{drop}} = 30\%$ for hidden layers and no dropout at the input layer. Hyperparameter search was performed by systematically exploring various combinations of learning rates, decay rates, thresholds and dropout rates, while leaving the values for time steps and neuron counts fixed.

A classification is deemed correct, if the output neuron, which corresponds to the target class, fires the strict maximum number of spikes. That is, all other output neurons fire fewer spikes. We plotted the test accuracy against the number of neurons $N$ and time steps $T$ as seen in Figures \ref{fig: plot architecture} and \ref{fig: plot timesteps}, respectively. A higher number of neurons increases the ability of the network to recognize details in the input samples. As expected, the classification accuracy of our network tends to increase with the neuron count. Saturation occurs when additional neurons do not lead to a considerable increase in accuracy. The most efficient network structure contains $2*800 + 800 = 2400$ hidden neurons, with $N=800$. It reaches a test accuracy of 98.41\%. The highest accuracy of 98.48\% can be reached at the cost of increase in neurons. With regards to Figure~\ref{fig: plot timesteps}, the hyperparameter tuning was performed for $T = 20$, explaining the peak accuracy for this configuration. With spike-based learning, every time step influences the update of weights. Although longer spike trains can represent more information, it also requires additional adjustments of learning rate and thresholds. Using only five time steps causes significant information loss in the activations, which is reflected in the drop of classification accuracy.

For a training with parameters $N = 800$ and $T = 20$, we analyze the learning rule's impact on the neurons of the network. After each epoch, we evaluate the network with the test dataset and record the local loss of each neuron in the three layers according to Equation~\ref{eqn: desire loss}. The result is plotted in Figure~\ref{fig: local errors}, which shows a clear decrease in local losses throughout the training process. Concurrently, the classification accuracy tends to increase. A deeper insight in the spiking activity of individual neurons for the first five training epochs is given in Figure~\ref{fig: spike activity}, with special regions of interest marked by green circles. We randomly selected 20 neurons from the two hidden layers and, together with the ten output neurons, plot their color-coded relative spike count. The activity is measured separately for each of the ten classes of digits 0 to 9. The last row of plots represents the output layer and, as expected, the spike activity is concentrated at the correct output neuron for each class, i.e. neuron 0 spikes for class 0, etc. One can observe that the activity of target neurons is initially lower and increases throughout the training (region~1). Accordingly, non-target output neurons tends transition towards lower activity (region~2). In the hidden layers, represented by the upper two rows of plots, neurons are shown to be sensitive to certain input classes. For example, neuron 11 in the first layer spikes for class 6, but not for any other class (region~3). This sensitivity is established mainly within the first training epoch. Bright regions at the left edges of some plots show that, initially, many neurons were active and later silenced as result of the training (region~4). Lastly, in Figure~\ref{fig: contribution}, colors encode how a neuron contributes on average to satisfy the desire of the neurons in the next layer. Contribution is a function of layer weights, input/output spikes, and desire values. Thereby, red represents a bad influence, meaning that spike activity in this neuron will cause neurons in the next layer to fire in an undesired manner. Blue color reflects positive contribution to the overall desire satisfaction. The contribution was measured for each of the 784 neurons of the first layer over the first 1000 training samples of the first epoch. Because neurons at the edges of the input image remain silent for all samples, their contribution is zero throughout. The others display negative contribution at the start of the training, gradually improving as the network is exposed to more training samples. The red patch at around 500 samples indicate a temporary drop in contribution, likely caused by out-of-distribution training samples.

Table~\ref{tbl: results mnist} compares our desire backpropagation with state-of-the-art works for SNN training. For a fair comparison, we include learning rules, which perform supervised spike-timing-driven weight updates, and employ fully-connected layers. Our method outperforms most of those learning algorithms in terms of accuracy, such as~\cite{shrestha2017stable, neftci2017event, kheradpisheh2018stdp, tavanaei2019bp, shrestha2019approximating, hao2020biologically, comsa2020temporal, liu2021sstdp, mirsadeghi2021stidi, tang2021biograd, luo2022supervised}. Those achieving a higher accuracy~\cite{zhang2018plasticity, falez2019multi} require a significant increase in the number of neurons and/or time steps. CNN architectures~\cite{kheradpisheh2018stdp, falez2019multi} especially require considerably more computational effort, with only a small impact on accuracy. While~\cite{neftci2017event, shrestha2019approximating, tang2021biograd} show advantages in terms of neuron count, they require feedback weights in addition to the feed-forward weights, which might not be suitable for memory-constraint devices. While all listed learning rules are able to train at least the last network layer in a supervised, some learning rules cannot be generalized for deeper networks~\cite{tavanaei2019bp, falez2019multi, hao2020biologically}. The table shows that desire backpropagation is a competitive learning rule, which achieves high accuracy while keeping the number of neurons low. 

\begin{table}[ht]
    \caption{Comparison of state-of-the-art learning rules for the MNIST dataset with regards to accuracy (in \%), number of hidden neurons and time steps.}
    \normalsize
    \renewcommand{\arraystretch}{1.1}
    \begin{center}
    \begin{tabular}{l|l l l|r r r}
        \hline
        \textbf{Model} & \textbf{Layers} & \textbf{Learning Rule} & \textbf{Acc.} & \textbf{Hid. Neurons} & \textbf{T.steps} \\
        \hline
        Shrestha (2017) \cite{shrestha2017stable}        & 2 FC & Stable STDP      & 89.7 &  1600 &  200 \\
        Neftci (2017) \cite{neftci2017event}             & 3 FC & eRBP             & 98.0 &  1000 & ~125 \\
        Kheradpisheh (2018) \cite{kheradpisheh2018stdp}  & CNN  & STDP+SVM         & 98.4 & 12220 &   30 \\
        Zhang (2018) \cite{zhang2018plasticity}          & 2 FC & Equ. Learn.+STDP & 98.5 &  4500 &  100 \\
        Tavanaei (2019) \cite{tavanaei2019bp}            & 3 FC & BP-STDP          & 97.2 &   650 &  ~10 \\
        Falez (2019) \cite{falez2019multi}               & CNN  & STDP+SVM         & 98.6 & 35328 &   -- \\
        Shrestha (2019) \cite{shrestha2019approximating} & 3 FC & EMSTDP           & 97.3 &  1000 &  200 \\
        Hao (2020) \cite{hao2020biologically}            & 2 FC & DA-STDP          & 96.7 & 10000 &  700 \\
        Comsa (2020) \cite{comsa2020temporal}            & 2 FC & Error BP         & 98.0 &   340 &   10 \\
        Mirsadeghi (2021) \cite{mirsadeghi2021stidi}     & 2 FC & STiDi-BP         & 97.4 &   500 & ~300 \\
        Liu (2021) \cite{liu2021sstdp}                   & 2 FC & SSTDP            & 98.1 &   300 &   16 \\
        Tang (2021) \cite{tang2021biograd}               & 3 FC & BioGrad          & 98.1 &   600 &   20 \\
        Luo (2022) \cite{luo2022supervised}              & 2 FC & Multi FE-Learn   & 98.1 &   800 &   30 \\
        \textbf{This work}                               & 3 FC & Desire BP+STDP   & 98.4 &  2400 &   20 \\
        \hline
    \end{tabular}
    \label{tbl: results mnist}
    \end{center}
\end{table}

\subsection{Fashion-MNIST classification}
Just like MNIST, Fashion-MNIST~\cite{xiao2017fashion} consists of 60,000 training and 10,000 validation images that are 28x28 pixels. They belong to ten kinds of clothing, such as shirts, trousers, and shoes. The photographic nature of the images contains more detail than digits of the MNIST dataset and are hence more challenging to classify correctly. However, for the same reason, Fashion-MNIST is considered a more realistic dataset for computer vision applications~\cite{hao2020biologically}. We train a network with three fully-connected layers and 784 -- 1000 -- 100 -- 10 neurons. Spike trains have a length of $T = 20$. A total of 600 epochs had to be computed due to higher dropout rates of 40\% and 5\% for hidden and input layers, respectively. All other hyperparameters are equal to the ones used for MNIST classification in Section \ref{sec: MNIST classification}.

With those settings, we achieved a classification accuracy of 87.56\%. The comparison with Shrestha~et~al. \cite{shrestha2019approximating} and Hao~et~al.~\cite{hao2020biologically} reflects the results of the MNIST classification (see Table \ref{tbl: results fashion mnist}). Desire backpropagation achieved the highest accuracy, while using either a similar number of neurons or less.

\begin{table}[ht]
    \caption{Results and comparison of Fashion-MNIST classification with indication of accuracy (in \%), number of hidden neurons and time steps.}
    \normalsize
    \renewcommand{\arraystretch}{1.1}
    \begin{center}
    \begin{tabular}{l|l l|r r r}
        \hline
        \textbf{Model} & \textbf{Layers} & \textbf{Learning Rule} & \textbf{Acc.} & \textbf{Neurons} & \textbf{T.steps} \\
        \hline
        Shrestha (2019) \cite{shrestha2019approximating} & 3 FC & EMSTDP         & 86.1 & 1000 & 200 \\
        Hao (2020) \cite{hao2020biologically}            & 2 FC & DA-STDP        & 85.3 & 6400 &  -- \\
        \textbf{This work}                               & 3 FC & STDP+Desire BP & 87.6 & 1100 &  20 \\
        \hline
    \end{tabular}
    \label{tbl: results fashion mnist}
    \end{center}
\end{table}

\subsection{Computational complexity}
\begin{figure}[t]
    \centering
    \begin{minipage}{0.45\textwidth}
        \centering
        \captionof{table}{Comparison of operations needed for three sub-processes using desire backpropagation and classical backpropagation. $I$/$O$ are the numbers of inputs/outputs to a neuron, $T$ is the number of time steps.}
        \normalsize
        \renewcommand{\arraystretch}{1.1}
        \setlength{\tabcolsep}{4pt}
        \begin{tabular}{l|l l l|l l}
            \hline
            \textbf{Process} & \multicolumn{3}{c|}{\textbf{SNN: Desire BP}} & \multicolumn{2}{c}{\textbf{ANN: BP}} \\
            & Mult. & Add. & Cond. & Mult. & Add. \\
            \hline
            Forward  & $1$ & $T I$ & $T I$ & $I$  & $I$ \\
            Backward & $O$ & $2O$  & $1$   & $2O$ & $O$ \\
            Update   & $1$ & --    & $1$   & $2$  & --  \\
            \hline
        \end{tabular}
        \label{tbl: complexity}
    \end{minipage}
    \hfill
    \begin{minipage}{0.52\textwidth}
        \centering
        \includegraphics[width=\textwidth]{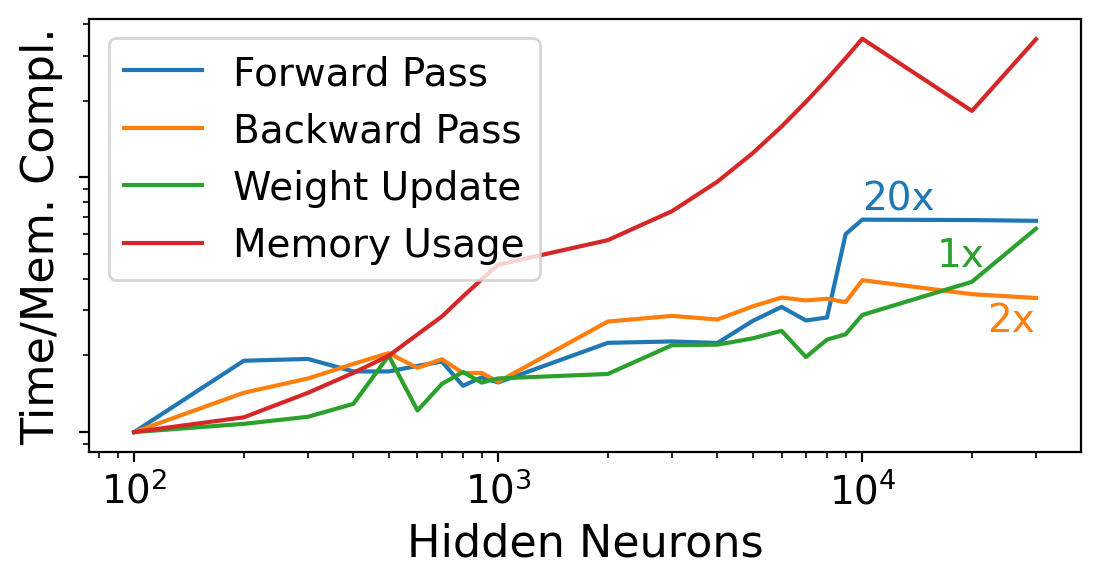}
        \caption{Normalized execution time of forward pass, backward pass, and weight update, and memory usage during training. Numbers on plot lines are the scaling factors to indicate relationship between execution times.}
        \label{fig: complexity}
    \end{minipage}
\end{figure}

Spiking neural networks have been touted as an efficient alternative to conventional artificial neural networks. In this section, we take a deeper look into the computational complexity of desire backpropagation. For forward pass, backward pass and weight update of a single neuron, the number of operations is determined and compared with classical backpropagation in Table~\ref{tbl: complexity}. It shows the dependency of the SNN forward pass on the number of time steps $T$ and the number of inputs $I$. Binary spikes allow the use of conditional statements instead of multiplications. Only one multiplication is needed for the decay of the membrane potential, whereas ANNs require multiplications and additions for all $I$ inputs. In the backward pass, Desire BP replaces one multiplication with an addition due to the local computation of losses. One conditional statement is used in the ternarization function. The ternary nature of the desire values benefits the weight update, as one multiplication operation can be saved. Multiplications are computationally more expensive than additions, which are in turn more expensive than conditional statements. Spike-based processing in combination with our learning rule can therefore be expected to be more computationally efficient overall.

From table~\ref{tbl: complexity} we can derive linear time complexity with respect to inputs $I$ and outputs $O$. To confirm this hypothesis, an SNN for the classification of MNIST digits was constructed, consisting of input, hidden, and output layer. The number of neurons in the hidden layer is varied in logarithmic steps from 100 to 30000. We then executed the training on an NVIDIA GeForce RTX 2080 Ti GPU and measured the execution time of forward pass, backward pass and weight update. In addition, we recorded the memory utilization of the process running the Python script.

Before displaying the results in Figure~\ref{fig: complexity}, the data were individually normalized in terms of offset and range, allowing a direct comparison of the dependency on the hidden neuron number. The plot uses logarithmic scale for both horizontal and vertical axes. As expected, a roughly linear trend can be detected for time complexity. Because each neuron in the network comprises an equal number of internal variables, the memory shows linear behaviour as well. Due to parallel execution within the GPU, execution time does not always increase with larger neuron count. Instead, it maintains a constant level before increasing by a bigger step. The absolute relationship between the execution time for different processes is indicated by the scaling factors on the plot lines. The forward pass takes around $20\times$ more time to run. With the number of time steps being 20, the forward pass has to execute 20 times, while backward pass and weight update is only performed once.

\section{Conclusion}
In this paper, we propose desire backpropagation, a novel lightweight learning algorithm for spiking neural networks. With STDP as its foundation, our learning rule utilizes the benefits of spike-based based learning such as resource efficiency and biological plausibility. To enable multi-layer learning, we introduced the ternary spiking desire of a neuron. It determines, whether the weight needs to be increased or decreased during the STDP update to minimize the output error. The desire value of the output layer can directly be derived from the labels of the training samples. For hidden layers, we utilized local losses, which can be calculated based on already propagated desire values. The backpropagation sets desire values in a way that helps satisfying the desire of the successive layer. The ternary desire values are coupled with a binary dropout mask to avoid overfitting of the model.

We tested our learning rule on the MNIST and Fashion-MNIST datasets and achieved remarkable classification accuracy of 98.41\% and 87.56\%, respectively. It not only shows superior accuracy compared to many other learning rules, but also uses less neurons and time steps than most other algorithms. We further compared the required operations with classical backpropagation and demonstrated its advantages in terms of computational complexity. Its performance and efficiency makes desire backpropagation a potential candidate for deployment on resource-constrained devices.

\section{Acknowledgement}
This work was supported by the Singapore Government’s Research, Innovation and Enterprise 2020 Plan (Advanced Manufacturing and Engineering domain) under Grant A1687b0033.

\bibliography{main}

\end{document}